\documentclass[10pt, sigconf, nonacm]{acmart}
\usepackage[utf8]{inputenc}
\usepackage{tikz}
\usepackage{listings}
\usepackage{amsmath}
\usepackage{lipsum}
\usepackage{mathtools}
\usepackage{cuted}
\usepackage{pifont}
\usepackage{amsfonts}
\usepackage{relsize}
\usepackage{booktabs}
\usepackage[noend]{algpseudocode}
\usepackage{algorithm}
\usepackage{afterpage}
\usepackage{tablefootnote}
\usepackage{xspace}
\usepackage[most]{tcolorbox}

\newcommand{\namex}{InfAdapter\xspace}

\title{\huge{Reconciling High Accuracy, Cost-Efficiency, and Low Latency\\of Inference Serving Systems}}

\author{Mehran Salmani$^*$, Saeid Ghafouri$^{\mathsection}$$^{\ddagger}$, Alireza Sanaee$^{\mathsection}$, Kamran Razavi$^{\dagger}$, \\Max M\"uhlh\"auser$^{\dagger}$, Joseph Doyle$^{\mathsection}$, Pooyan Jamshidi$^{\ddagger}$, Mohsen Sharifi$^*$} {
\affiliation{
 \institution{Iran University of Science and Technology$^*$, Queen Mary University of London$^{\mathsection}$, \\Technical University of Darmstadt$^{\dagger}$, University of South Carolina$^{\ddagger}$}
}

}

\begin{document}

\acmYear{2023}\copyrightyear{2023}
\acmConference[EuroMLSys '23]{3rd Workshop on Machine Learning and Systems}{May 8, 2023}{Rome, Italy}
\acmBooktitle{3rd Workshop on Machine Learning and Systems (EuroMLSys '23), May 8, 2023, Rome, Italy}
\acmPrice{15.00}
\acmDOI{10.1145/3578356.3592578}
\acmISBN{979-8-4007-0084-2/23/05}




\begin{abstract}
The use of machine learning (ML) inference for various applications is growing drastically. ML inference services engage with users directly, requiring fast and accurate responses. Moreover, these services face dynamic workloads of requests, imposing changes in their computing resources. Failing to right-size computing resources results in either latency service level objectives (SLOs) violations or wasted computing resources. Adapting to dynamic workloads considering all the pillars of accuracy, latency, and resource cost is challenging. In response to these challenges, we propose \namex, which proactively selects a set of ML model variants with their resource allocations to meet latency SLO while maximizing an objective function composed of accuracy and cost. \namex decreases SLO violation and costs up to 65\% and 33\%, respectively, compared to a popular industry autoscaler (Kubernetes Vertical Pod Autoscaler).
\end{abstract}

\maketitle
\thispagestyle{plain}
\pagestyle{plain}
\vspace{-1em}
\section{Introduction}
\label{sec:intro}


The computing demand for machine learning (ML) has exponentially increased over the past decade \cite{amodei-hernandez-2019}. For example, different ML applications, including computer vision, machine translation, chatbots, medical, and recommender systems, are running in data centers \cite{bar2019, park2018deep, velikovich2018semantic, sarwinda2021deep}, comprising more than 90\% of computing resources allocated to ML~\cite{bar2019, leopold2019, akoush2022desiderata}. ML inference services are user-facing, which mandates high responsiveness \cite{zhang2017live, gujarati2017swayam}. Moreover, high accuracy is crucial for these services~\cite{nigade2022jellyfish, gunasekaran2022cocktail}. Consequently, inference systems must deliver highly accurate predictions with fewer computing resources (cost-efficient) while meeting latency constraints under workload variations~\cite{zhang2017live, razavi2022fa2, gujarati2017swayam, gujarati2020serving}.

\begin{table}[t!]
\caption{
\namex is superior compared to the state-of-the-art solutions. ($\ast$) Cocktail uses model ensembling leading to cost inefficiencies in particular scenarios (see Section \ref{sec:relatedwork}).}
\begin{tabular}{|l|c|c|c|c|c|}
\hline
 \textbf{Feature} & {\rotatebox[origin=c]{90}{MS~\cite{zhang2020model}}} & {\rotatebox[origin=c]{90}{INFaaS~\cite{romero2021infaas}}} & {\rotatebox[origin=c]{90}{Cocktail~\cite{gunasekaran2022cocktail}}} & {\rotatebox[origin=c]{90}{VPA~\cite{vpa}}} & {\rotatebox[origin=c]{90}{\namex}} \\ \hline

Cost Optimization & \textcolor{red}{\ding{53}}  & \textcolor{teal}{\ding{51}} & \textcolor{teal}{\ding{51}}$\ast$ & \textcolor{teal}{\ding{51}} & \textcolor{teal}{\ding{51}} \\ \hline
Accuracy Maximization & \textcolor{teal}{\ding{51}} & \textcolor{red}{\ding{53}} & \textcolor{teal}{\ding{51}} & \textcolor{red}{\ding{53}} & \textcolor{teal}{\ding{51}} \\ \hline
Predictive Decision-Making & \textcolor{teal}{\ding{51}} & \textcolor{red}{\ding{53}} & \textcolor{teal}{\ding{51}} & \textcolor{teal}{\ding{51}} & \textcolor{teal}{\ding{51}} \\ \hline
Container as a Service (CaaS) & \textcolor{red}{\ding{53}} & \textcolor{red}{\ding{53}} & \textcolor{red}{\ding{53}} & \textcolor{teal}{\ding{51}} & \textcolor{teal}{\ding{51}} \\ \hline
Latency SLO-aware & \textcolor{teal}{\ding{51}} & \textcolor{teal}{\ding{51}} & \textcolor{teal}{\ding{51}} & \textcolor{red}{\ding{53}} & \textcolor{teal}{\ding{51}} \\ \hline
\end{tabular}
\label{table:comparison}
\end{table}

The dynamic nature of inference serving workloads requires different resource allocations for ML services~\cite{zhang2017live, gujarati2017swayam}. Failing to right-size the services results in over or under-resource provisioning. Under-provisioning leads to service level objective (SLO) violations~(\emph{e.g.,} $99^{th}$ percentile of latency distribution, P99-latency)\cite{gujarati2017swayam, zhang2019mark}. Conversely, over-provisioning wastes computing resources~\cite{romero2021infaas, zhang2019mark}. To address these problems caused by dynamic workloads, \textsc{Auto-scaling} \cite{gujarati2017swayam, zhang2019mark, vpa, hpa, romero2021infaas, gandhi2012autoscale} resizes the resources of the service, and \textsc{Model-switching} \cite{zhang2020model, nigade2022jellyfish} switches between ML model variants that differ in their inference latency and accuracy (higher accuracy, higher latency); the former tries to be cost-efficient, and the latter tries to be more accurate, while both guarantee latency SLOs.

\textsc{Auto-scaling} and \textsc{model-switching} as the state-of-the-art adaptation mechanisms fail to consider the accuracy and cost-efficiency simultaneously. \textsc{Auto-scaling} may sacrifice accuracy if it works with a low-accuracy model variant or incur high resource costs if used for a high-accuracy model variant. Conversely, \textsc{model-switching} can be a subject of under-provisioning in cases where even the least accurate model variant cannot respond to the workload; it also fails to be cost-efficient when the capacity of the most accurate model variant is more than the workload on the service.


The ability to jointly resize and switch ML model variants provides new opportunities. For instance, our experiments demonstrate that a Resnet50 model variant on 8 CPU cores allocation can sustain almost the same load that a Resnet152 variant does with 20 CPU cores; moreover, a Resnet18 with 8 CPU cores can process the same load as a Resnet50 with 20 cores, while meeting P99-latency (750ms). Using a set of model variants instead of a single variant provides more granular accuracy/cost trade-offs.

Inference systems should be adaptive in response to dynamic workloads and be able to consider all the \textit{contrasting} objectives, including responsiveness~(latency), accuracy, and cost-efficiency (allocated CPU cores) when dedicating resources to services and models. Moreover, reconciling these three measures is challenging as achieving one causes a violation or sacrifice of the other, and finding a trade-off among these three is daunting.

In response to these challenges, we design \namex and empirically show that it can address the limitations of existing solutions. It predicts the service workload to mitigate provisioning overhead and, by using the predicted load, selects a set of ML model variants and their sizes (CPU cores) as the service backends to meet latency SLO and to maximize an objective function composed of average accuracy, resource cost, and loading cost (Section~\ref{sec:formulation}). To process the incoming requests, we implemented a dispatcher that load balances user requests to the backend variants according to their capacity(Section~\ref{sec:methodolgy}). Our experiments demonstrate that \namex reduces average accuracy loss for latency SLO of 750~ms at $99^{th}$ percentile up to 4\% given the same load compared to existing solutions (Section~\ref{sec:results}).

%


\begin{figure}[!t]
\centering
\includegraphics[scale=0.5]{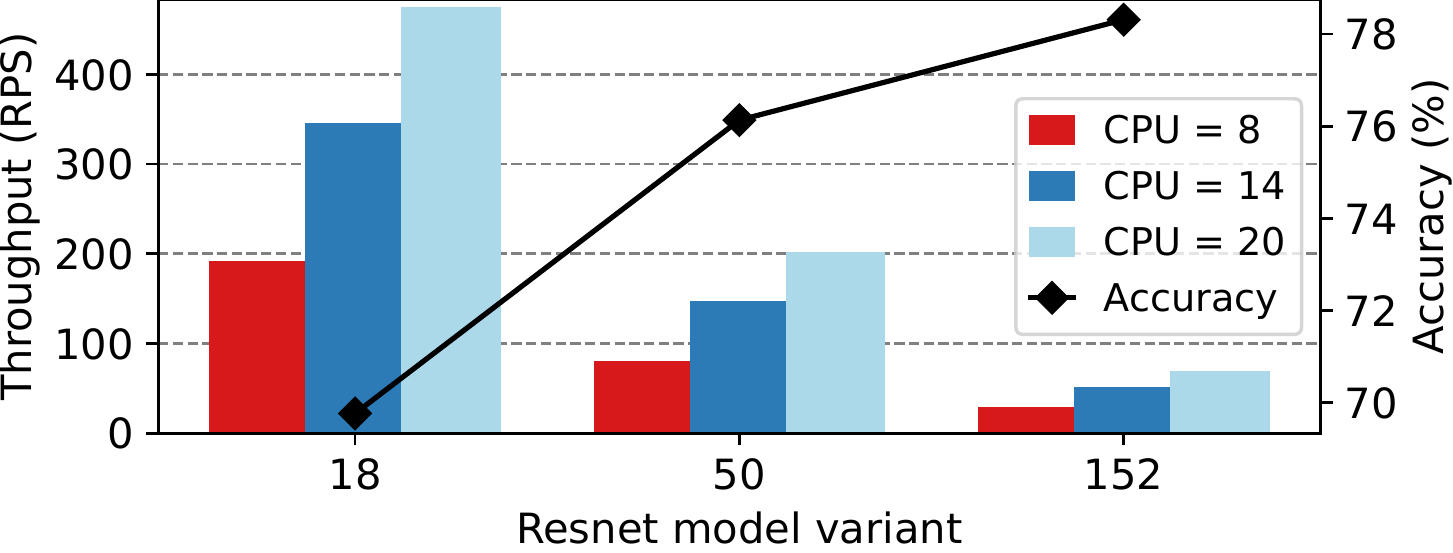}
\caption{Throughput of three Resnet variants under 8, 14, and 20 CPU cores. We ensured the latency of all configurations is lower than 750~ms at the P99-latency at the saturation load.}

\label{fig:cpu-arch}
\end{figure}


We have prototyped \namex\footnote{https://github.com/reconfigurable-ml-pipeline/InfAdapter} in a Kubernetes cluster and used TensorFlow Serving \cite{olston2017tensorflow} model server to serve our ML models.
We experimentally evaluate \namex using a real workload trace, Twitter-trace~\cite{twitter-trace-2021-08} (Section~\ref{sec:results}) and compare it against existing solutions (\emph{e.g.,} vertical Pod auto-scaler~(VPA)~\cite{vpa}, and Model-Switching~\cite{zhang2020model}). Our experiments illustrate that \namex reduces SLO violations by up to 65\% compared to existing solutions. We further open-sourced our implementation for community engagement and reproducibility of our experiments. Table \ref{table:comparison} summarizes differences between \namex and other existing approaches.






\definecolor{mygray}{RGB}{220,220,220}



\section{Motivation}

Due to interaction with online users, inference services are latency-sensitive \cite{gujarati2017swayam, zhang2019mark}, and since they contain heavy computations, they are resource-intensive \cite{akoush2022desiderata}. Accuracy is also a pillar dimension of these services~\cite{nigade2022jellyfish}. Faced with dynamic workload~\cite{zhang2017live, gujarati2017swayam}, it is essential to consider the ternary trade-off space between latency, accuracy, and the resource cost dynamically to address latency requirements while gaining higher accuracy cost-efficiently.


The importance of having accurate predictions while being cost-efficient, on one hand, hinders us from selecting a computationally light model variant with a low accuracy; on the other hand, selecting the most accurate model requires a very high resource cost to fulfill latency SLOs, which may even be unavailable. We conducted experiments with Resnet model variants under different CPU core assignments and captured their sustained throughput (number of requests they could handle given 750ms P99-latency SLO). Figure~\ref{fig:cpu-arch} shows our experiment's result for Resnet18, Resnet50, and Resnet152 variants used for image classification under 8, 14, and 20 CPU core assignments. In the figure, a Resent18 with 8 CPU cores, and a Resnet50 with 20 CPU cores, can almost sustain the same throughput under the latency SLO; a similar argument is applicable to Resnet50 with 8 cores and Resnet152 with 20 cores. Due to the fact that latency-accuracy trade-off space changes based on the workload, it is a non-trivial task to pick the right model variant with the right resource allocation.

\begin{tcolorbox}[colback=green!5!white,colframe=green!75!black]
  \textbf{Observation 1}. ML model variants provide the opportunity to reduce resource costs while adapting to the dynamic workload.
\end{tcolorbox}
 
\begin{figure}[t!]
\centering
\includegraphics[scale=0.5]{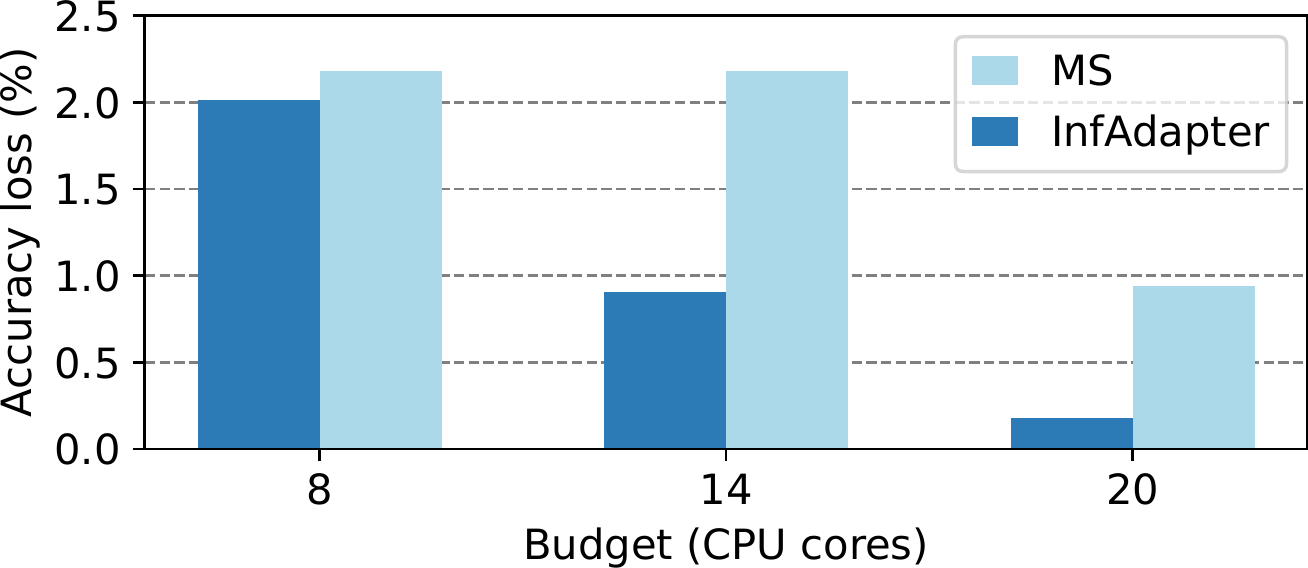}
\caption{Comparison of \namex (the ability to use a set of alternative models in the back-end) and Model-Switching+ on accuracy loss (accuracy of the most accurate model, Resnet152, subtracted by the accuracy of each bar). Each bar sustains the SLO of 750ms at P99-latency for a 75RPS load.}
\vspace{-1.3em}
\label{fig:infadapter-solution-space}
\end{figure}

Using the traces collected from the previous experiment, we used two approaches to select backend model variant(s) to sustain a 75RPS load under a 750ms P99-latency SLO, using different CPU budgets (8, 14, and 20 CPU cores). Approach 1 is to opportunistically take advantage of selecting both a set of model variants and also their sizes (\namex). Approach 2 is to select only one model variant and its size (MS). For example, under a 14 CPU core budget, InfAdapter selects Resnet50, Resnet101, and Resnet152 with 2, 6, and 6 CPU cores, respectively. However, the most accurate model that MS can pick under the 14-core budget for 75RPS is Resnet50. We compared the two approaches' accuracy loss (i.e., the accuracy we obtained subtracted from the accuracy of our most accurate variant, Resnet152). In Figure~\ref{fig:infadapter-solution-space}, we observe that \namex is able to gain higher average accuracy (lower accuracy loss) for the requests by having more options to select from, i.e., selecting a set of models rather than a single model.


\begin{tcolorbox}[colback=green!5!white,colframe=green!75!black]
  \textbf{Observation 2}. Using a set of model variants simultaneously can provide better average accuracy compared to having one active variant.
\end{tcolorbox}

Given dynamic workloads, we propose an adaptive mechanism for ML inference services to achieve latency SLO-aware, highly accurate, and cost-efficient inference systems. \namex selects a subset of model variants to meet latency SLOs and maximizes an objective function of accuracy and cost. \namex reconciles three important yet contradictory objectives~(\emph{accuracy, cost, and latency}).

\section{Problem Formulation}

\begin{table}[!t]
    \centering
    \footnotesize
    \caption{Notations}\label{table:notation}
    \begin{tabular}{ p{1.2cm}p{5.8cm}  }
        \toprule
        \textbf{Symbol} & \textbf{Description}\\
        \midrule
        $M$ & Set of all model variants for a given task\\
        $L$ & Latency SLO\\
        $m$ & An ML model variant from set $M$\\
        $acc_m$ & Accuracy of variant $m$ \\
        $rt_m$ & Readiness time of variant $m$ \\
        $tc_m$ & Transition cost of variant $m$ \\
        $n_m$ & Number of CPU cores for variant $m$ \\
        $p_m(n_m)$ & Processing latency of variant $m$ with $n_m$ CPU cores\\
        $th_m(n_m)$ & Throughput of variant $m$ with $n_m$ CPU cores\\
        $AA$ & Average Accuracy\\
        $RC$ & Resource cost\\
        $LC$ & Loading cost\\
        $B$ & CPU budget\\
        $\lambda$ & Workload on the system\\
        $\lambda_m$ & Workload quota on variant $m$\\
        \bottomrule
    \end{tabular}
\end{table}

\label{sec:formulation}
We formally describe the accuracy-cost problem using ML model variants while guaranteeing the latency SLO.

We denote $M$ as the set of model variants for a specific task, with the latency SLO, $L$, the given accuracy of model $m \in M$, $acc_m$, and the model readiness time (loading the model into memory and the model initialization), $rt_m$. We profile our set of model variants under different CPU assignments to capture the number of requests they can process concerning latency SLO $L$. Furthermore, by using the profiled data, we train a linear regression model to estimate the processing latency and throughput of model variant $m \in M$ under any CPU cores $n_m \leq B$, $p_m(n_m), th_m(n_m)$, where $B$ is the total CPU budget and the total resource cost as $RC = \sum_{m \in M} n_m$.

To maintain system stability during a dynamic workload, the aggregated throughput of all available models for a given task must stay above an expected (predicted) request rate $\lambda$. Mathematically, this can be expressed as $\sum_{m \in M, n \leq B} th_m(n_m) \geq \lambda$. Moreover, we define the weighted average accuracy, based on the quota of the workload on variant $m$, $\lambda_m$, as $AA = \sum_{m \in M} \frac{\lambda_m}{\lambda} \cdot acc_m$. Furthermore, we define the model loading cost as $LC = \max \{ tc(m) * rt_m, \; m \in M \}$ where the transition cost, $tc(m)$, is equal to 1 if the model variant $m$ needs to be loaded, and 0 otherwise. Table~\ref{table:notation} summarizes the notations we use in the paper.


We define a multi-objective optimization problem to decide which subset of model variants to use such that under a given workload, the end-to-end latency is guaranteed. 
The goal is to maximize the weighted average accuracy $AA$ and to minimize the total resource $RC$ and loading $LC$ costs. The problem can be formulated with the following integer linear programming (ILP):








\begin{equation}
    \begin{aligned}
        \max &&& \alpha \cdot AA - (\beta \cdot RC + \gamma \cdot LC) \\
        \text{subject to} &&&  \lambda \leq \sum\limits_{m \in M} th_m(n_m), \\
        &&& \lambda_m \leq th_m(n_m)\\
        &&& p_m(n_m) \leq L, \forall m \in M, \\
        &&& RC \leq B, \\
        &&& n_m \in \mathbb{W}, \forall m \in M.
    \end{aligned}
    \label{eq:ip}
\end{equation}

In the objective function, we introduce $\alpha, \beta, \gamma$ to normalize the resource and loading costs and give importance to the objectives based on user preference. The first two constraints ensure the system's stability, e.g., there are enough resources to support the incoming workload. The third constraint satisfies the latency SLO, while the last two constraints bound the CPU core per model to be non-negative and within the available resources in the system. We use the Gurobi optimizer~\cite{gurobi} to solve the ILP in the above equation.

\section{System Design}
\label{sec:methodolgy}

\begin{figure}[t!]
\includegraphics[scale=0.31]{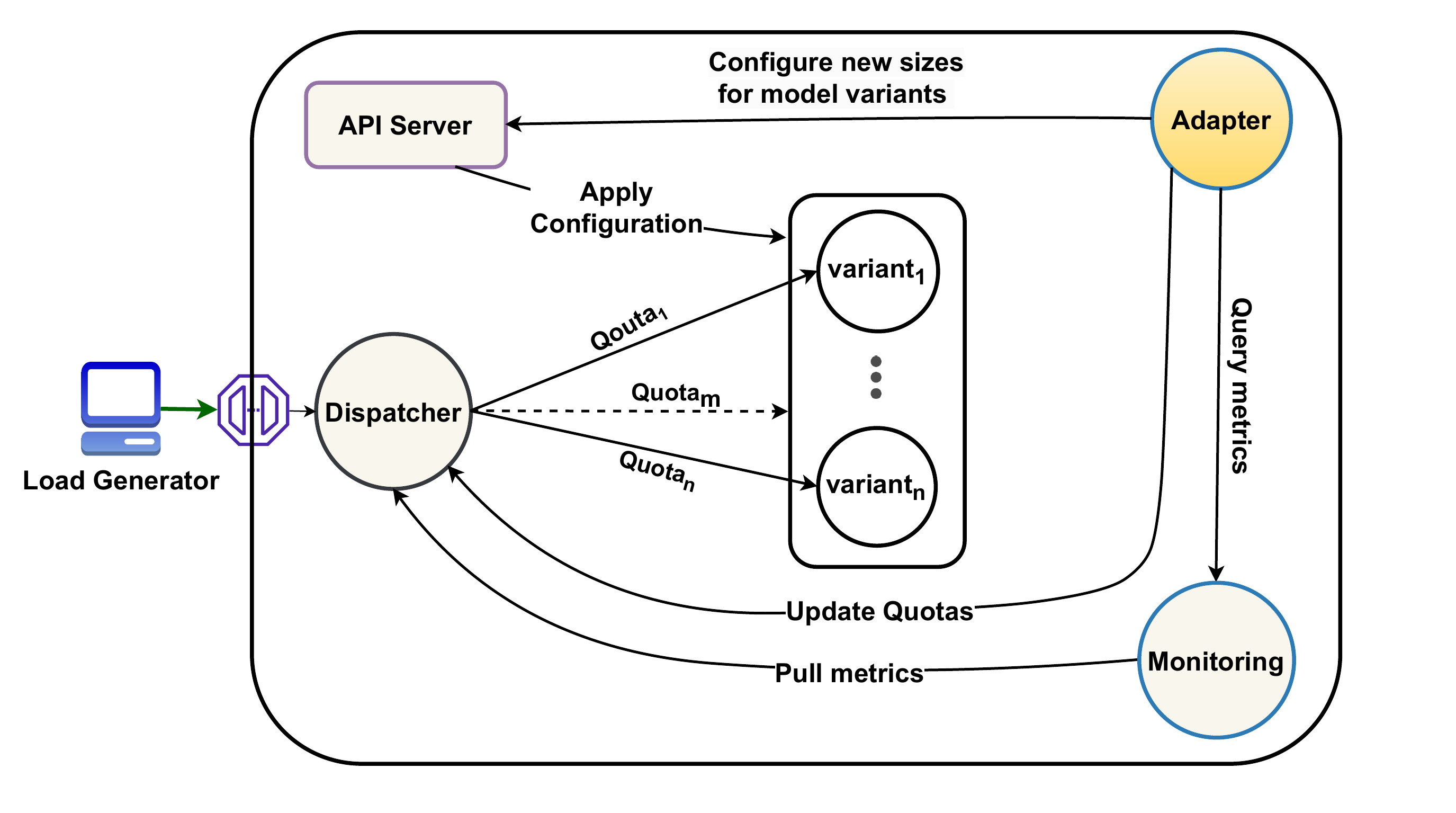}
\caption{\namex structure; variants can be scheduled (by the Kubernetes scheduler) in any of the nodes.}
\label{fig:project-structure}
\end{figure}


An overview of the \namex architecture is demonstrated in Figure~\ref{fig:project-structure}. The system consists of three major components (monitoring, adapter, and dispatcher). \emph{Monitoring} keeps monitoring statistics about the distribution of request arrivals. \emph{Adapter} is responsible for first predicting the next time-interval workload based on the workload history gathered from the monitoring component and then finding a set of model variants, their CPU cores, and their workload quota by solving the ILP in Equation~\ref{eq:ip}. \emph{Dispatcher} controls distributing the requests to the set of multi variants based on the models' workload quota provided by the \emph{Adapter} component.



\textbf{Monitoring.} The monitoring demon is in charge of fetching the arrival rate from the dispatcher. We get the number of requests per second and pass it to the forecaster to predict the arrival workload for the next time interval.

\textbf{Adapter.} The Adapter consists of two sub-components, a time-series forecaster and a solver. Time-series forecaster predicts future workload based on historical incoming workload patterns of requests on the system. The solver aims to solve the ILP in Equation~\ref{eq:ip} (every 30 seconds) to achieve the highest possible accuracy while respecting the latency SLO and available resources using the predicted workload and the current state of CPU allocation. Finally, the Adapter passes the set of models and their CPU cores to the cluster for enforcing the system configuration and the model's quota variables to the dispatcher for load balancing the incoming workload.


\textbf{Dispatcher.} The Dispatcher component load balances the incoming workload among the models in the cluster based on the weighted round-robin algorithm using the received models' quota variable, $\lambda_m$, from the solver in the adapter component. 



\section{Experimental Evaluation}
\label{sec:results}

\begin{figure}[t!]
\includegraphics[scale=0.58]{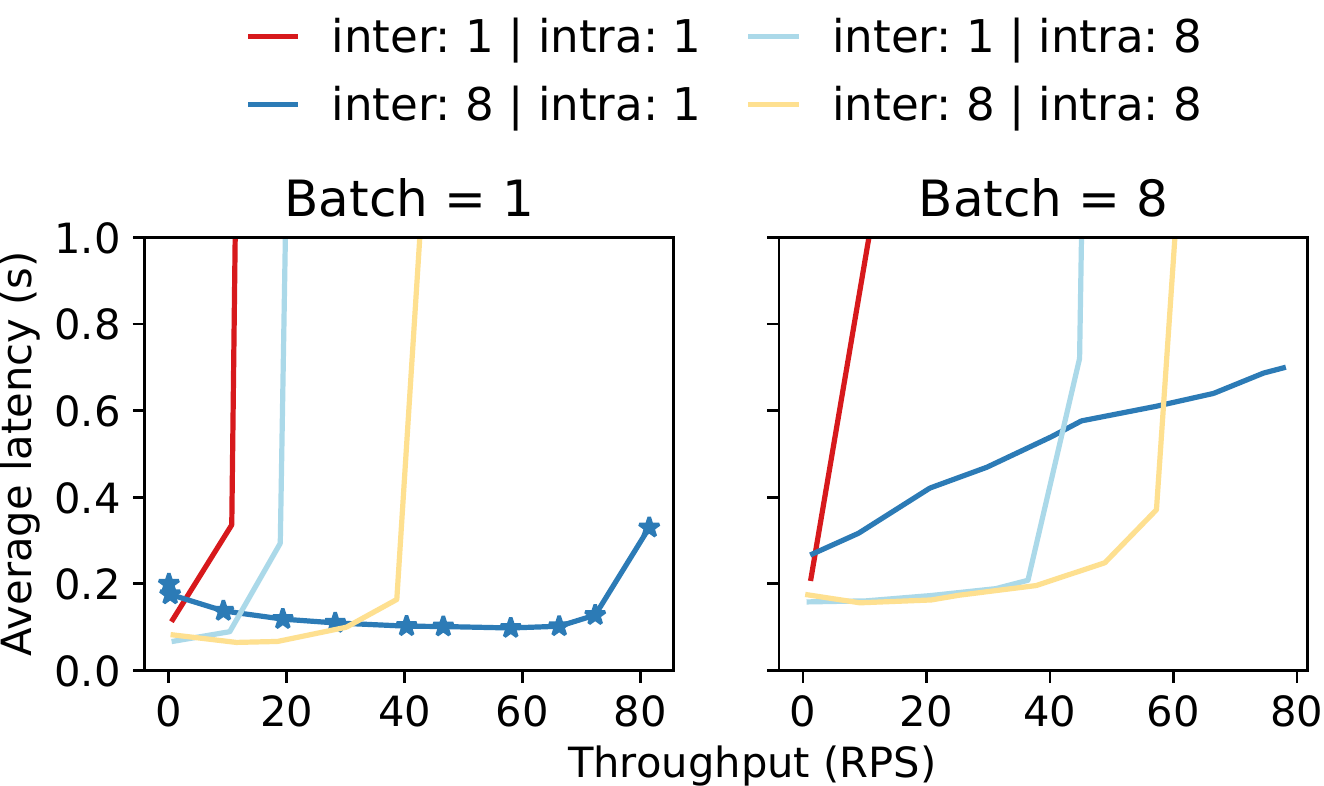}
\caption{Throughput-Average latency for batch sizes of 1 (batching disabled) and 8 with different parallelism configurations on Resnet50 with 8 CPU cores allocations. The starred configuration is the chosen configuration through our experiments.}
\label{fig:other-configs}
\end{figure}

\begin{figure}[t!]
\includegraphics[scale=0.55]{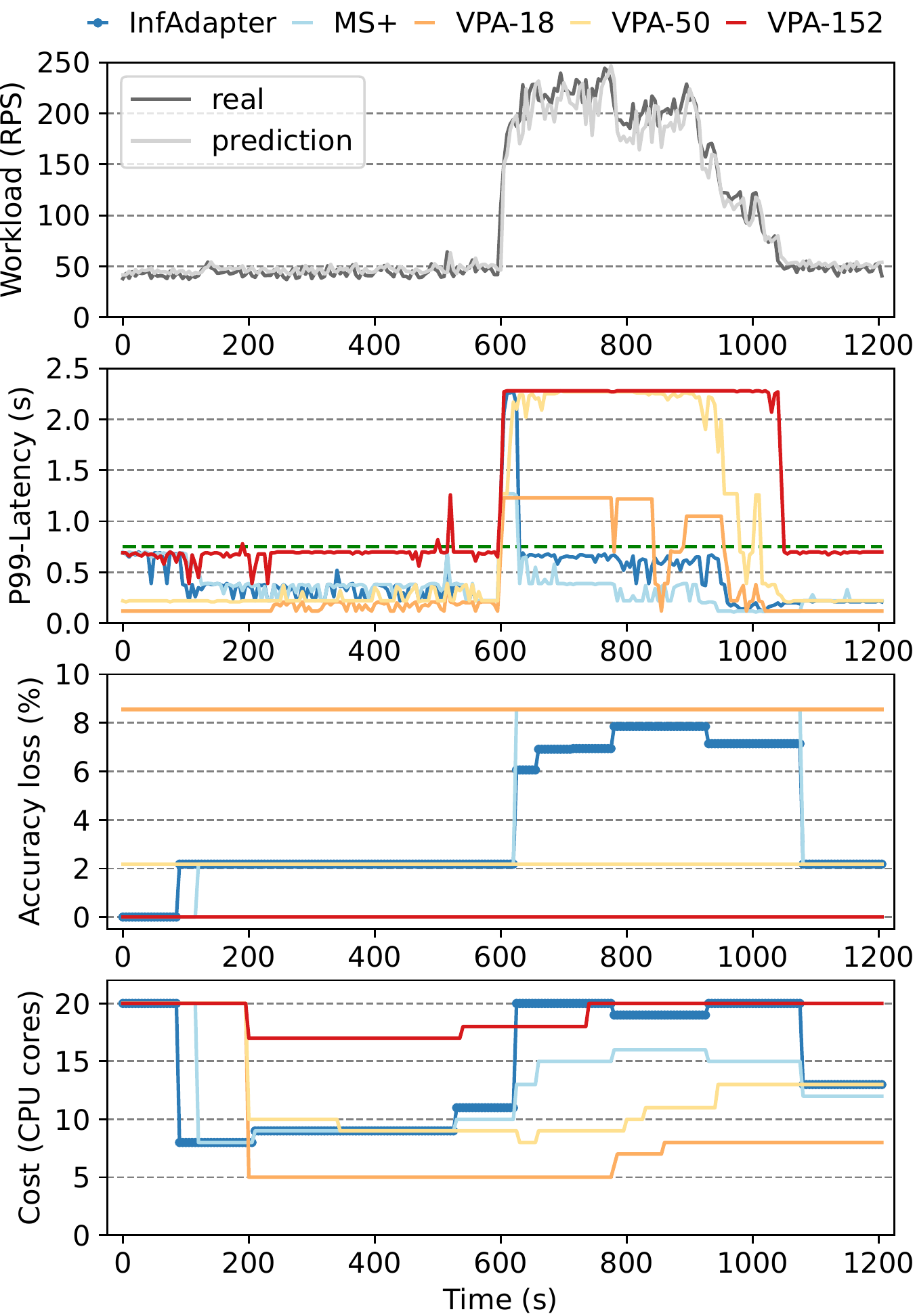}
\caption{Comparison of \namex with VPA used along with Resnet18, Resnet50 and Resnet152 on accuracy loss, cost and P99-latency, during the experiment, $\beta$ = 0.05.}
\vspace{-2em}
\label{fig:comparison-0.05}
\end{figure}


We have prototyped \namex in a Kubernetes cluster of two homogeneous physical machines from the Chameleon Cloud~\cite{keahey2020lessons}  equipped with 48 CPU cores of type Intel(R) Xeon(R) Gold 6126 CPU @ 2.60GHz and 192 GiB of RAM. TensorFlow Serving is used to serve the model variants in separate Docker containers.

\noindent \textbf{Batching and parallelism parameters.} Batching and parallelism parameters are practical configuration knobs of ML inference services. Batching refers to aggregating multiple requests into one request, which is widely adapted for GPU inference systems~\cite{crankshaw2017clipper, hu2021scrooge, shen2019nexus, wang2021morphling}. However, as shown in Figure \ref{fig:other-configs}, inference on CPU does not substantially benefit from batching in increasing the throughput, but increasing batch size leads to higher latency. Intra-op parallelism defines the parallelism degree within an operation~(such as matrix multiplication), and inter-op parallelism defines the parallelism across independent operations of inference requests~\cite{parallelism_torch, tf-inter, tf-intra}.


We measured the effect of batching and CPU intra/inter operation parallelism on Resnet50 with 8 CPU cores regarding throughput and latency. Experimental results are shown in Figure~\ref{fig:other-configs}, which captures throughput-average latency relation under different batching and parallelism configurations. We choose (starred plot) to disable batching (set to 1), set inter-op parallelism to the number of CPU cores, and disable intra-op parallelism (set to 1) in \namex across all the experiments to get the best throughput with a latency within the 750ms SLO. Further, we observed the same trend for all other model variants and CPU allocations. 

\begin{figure}[t!]
\includegraphics[scale=0.55]{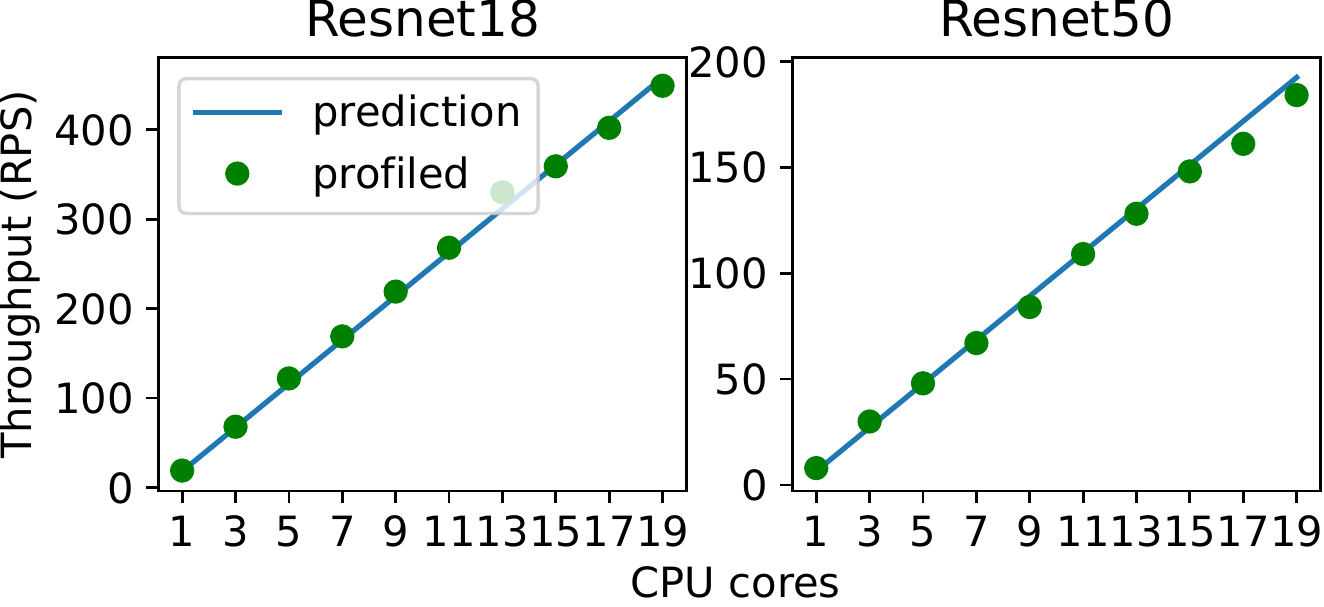}
\caption{Profiled values and predicted values of throughput of Resnet18 and Resnet50 models under different CPU allocations.}
\vspace{-1em}
\label{fig:regression-accuracy}
\end{figure}

\noindent \textbf{Profiling methodology} \namex optimization formulation (Equation~\ref{eq:ip}) needs to know the throughput $th_m (n_m)$ of each variant $m$ under different CPU core allocation $n_m$. We use a linear regression model for each ML model variant and train these regression models using the profiling data of only 5 CPU allocations (1, 2, 4, 8, and 16 cores) out of all possible allocations to avoid extra profiling costs. The regression models are then used to predict the throughput of the variants under any CPU allocation. Figure~\ref{fig:regression-accuracy} shows the prediction accuracy of regression models for Resnet18 and Resnet50. As observed in the experiments the predicted RPS can be accurately predicted using the regression model. For instance, the R-squared ($R^2$) for the regression models of Resnet18 and Resnet50 are $0.996$ and $0.994$ respectively.

\noindent \textbf{Load forecaster} We used LSTM~\cite{hochreiter1997long} for time-series forecasting. Our LSTM model takes as input the load per second of the past 10 minutes collected from the monitoring component, and predicts the maximum workload for the next minute; we used the first two weeks of Twitter-trace dataset~\cite{twitter-trace-2021-08} to train the LSTM model. The LSTM neural network comprises a 25-unit LSTM layer followed by a one-unit dense layer for the output layer. We used Adam optimizer with MSE loss function for training the network. Figure 5, the top plot, shows the prediction accuracy of the LSTM on a sample from the Twitter trace.

\noindent \textbf{\namex handles bursty and non-bursty workloads.} First, we experiment with bursty workloads to understand the performance of \namex. We compared \namex against an extended version of Kubernetes built-in Vertical Pod Autoscaler (VPA)~\cite{vpa} and an enhanced version of Model-Switching \cite{zhang2020model} (MS+). As the performance of the built-in VPA was very poor in the empirical evaluations, we made the following changes to it for a fair comparison against our approach. Initially, at each recommendation timestep, the built-in VPA removes the old container and then creates a new container with predicted resource allocations; this results in a service downtime during the recreation episode; to prevent this, we first create the container with the VPA recommended resources, and after it is up and running, remove the previous version. Secondly, we dropped the consideration of resource lower bound in VPA to scale up faster in response to the dynamic workload. For more information on the VPA algorithm, refer to \cite{rzadca2020autopilot}. Also, in MS+, since Model-Switching performs on a fixed resource budget, we add predictive allocation. At each time step, a model variant and its resource allocation are selected based on the same objective function we use for \namex in Equation~\ref{eq:ip}.

We evaluated the results on a 20-minute sample of Twitter-trace (Figure~\ref{fig:comparison-0.05} top) that contains a steady load (0-600s), a load spike (600s-800s), a gradual decrease in the load (800s-100s) and a sample of going back to the initial load (1000s-1200s). Almost all the compared methods can stay under the 750 ms SLO under a steady load. Once there is a load spike at 600s, almost all the compared methods suffer from SLO violations with a non-negligible margin (E.g., we observed a 10-minute violation for Resnet152). However, \namex and MS+ temporarily trade-off a little accuracy in favor of being responsive to the load spike with a short SLO violation time. Between MS+ and \namex, \namex can achieve the same SLO attainment with less accuracy loss during the load burst.

\begin{figure}[t!]
\includegraphics[scale=0.55]{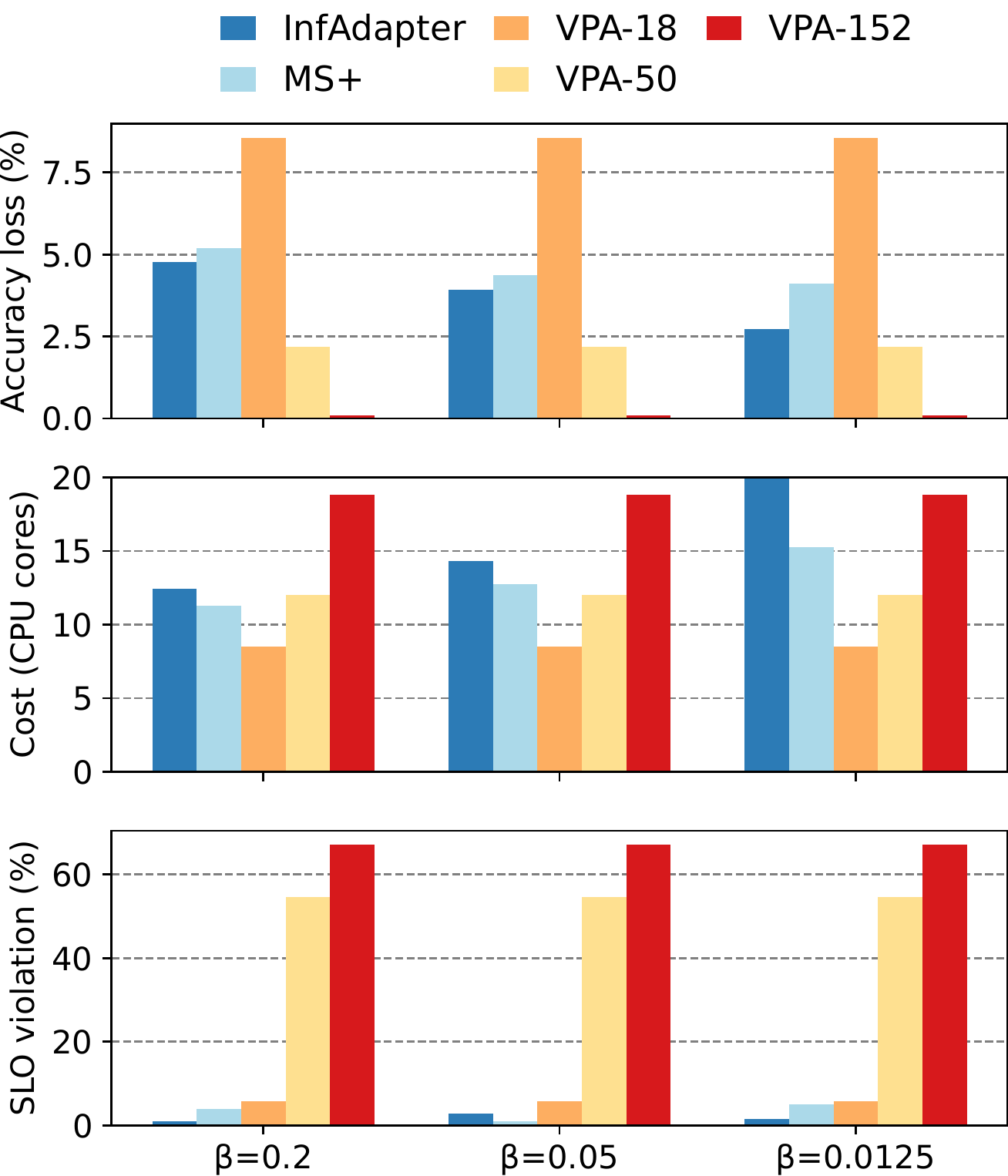}
\caption{Comparison of \namex with VPA used along with Resnet18, Resnet50 and Resnet152 on accuracy loss, cost and 99th percentile latency, for the whole experiment, under different $\beta$ values.}
\label{fig:comparison-agg-0.2}
\vspace{-1.25em}
\end{figure}

\namex aims to provide a tradeoff between accuracy and cost objectives. Under $\beta=0.05$, we observed that \namex could balance the cost and accuracy objectives and comply with latency SLO. The same trend can be identified from the cumulative result of the entire experiment in Figure~\ref{fig:comparison-agg-0.2}. The \namex can always balance the cost and accuracy objectives better than MS+. Also, VPA variants mostly took an extreme in maximizing only one objective; e.g., VPA-18 is the most cost-effective, but it comes at the expense of being very inefficient in accuracy.

\begin{figure}[t!]
\includegraphics[scale=0.55]{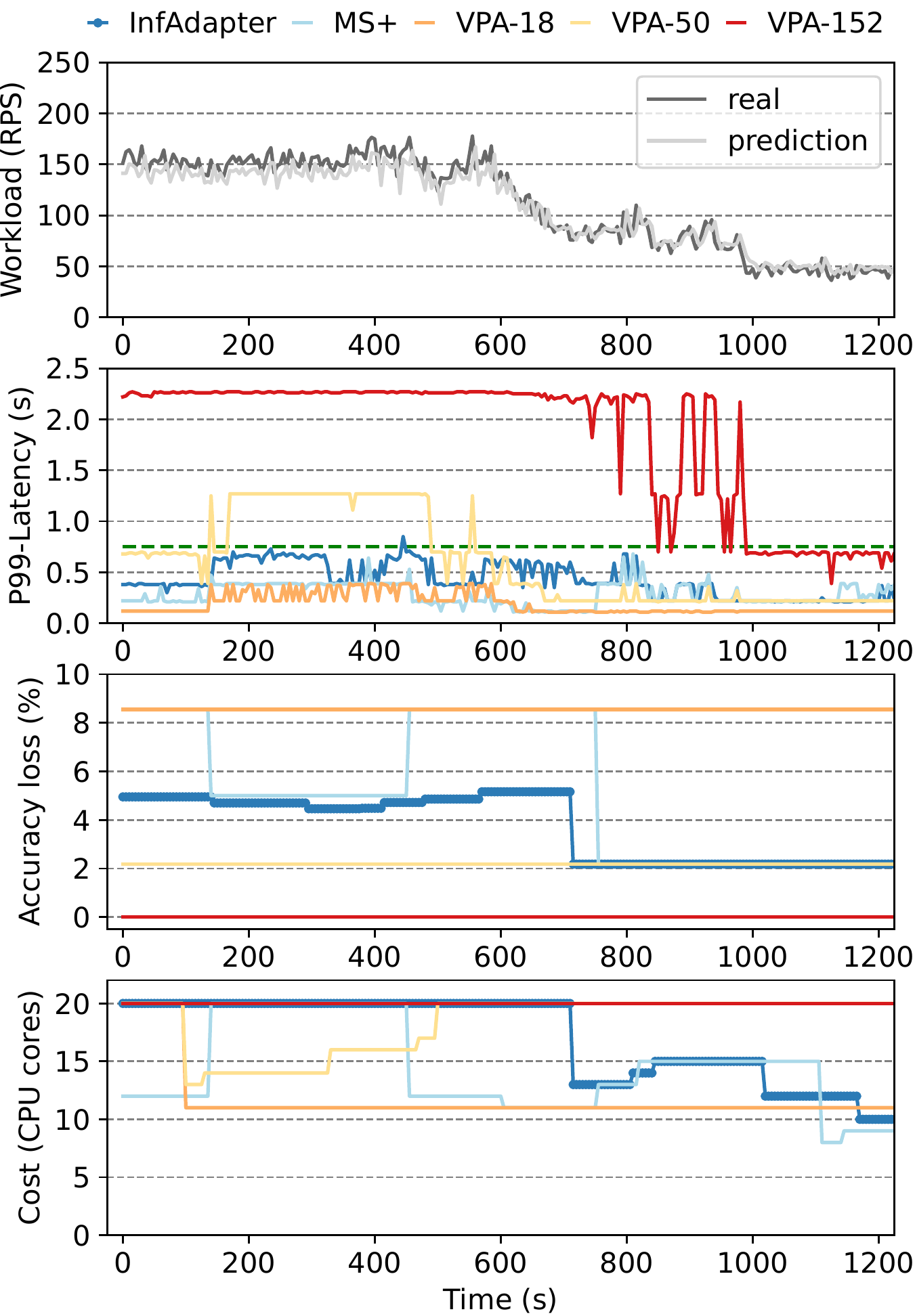}
\caption{Comparison of \namex with VPA used along with Resnet18, Resnet50 and Resnet152 on accuracy loss, cost and P99-latency, during the experiment with a non-bursty workload, $\beta$ = 0.05.}
\vspace{-1.25em}
\label{fig:comparison-0.05-nonbursty}
\end{figure}

Similarly, we used a non-bursty workload (Figure \ref{fig:comparison-0.05-nonbursty}). We observed that \namex has less accuracy loss than all other methods (except VPA+ with Resnet152, which has zero accuracy loss at the expense of high cost and SLO violations). Although in most cases \namex has better SLO compliance, the difference between MS+ and \namex in terms of cost and accuracy is small. We found that the difference was higher for a synthesized workload. In future work, we aim to evaluate \namex the cost and accuracy trade-off with respect to SLO guarantees with different workloads.

Refer to the appendix for experiments with $\beta=0.0125$ and $\beta=0.2$. We observed that lower values of $\beta$ \namex prioritize accuracy over resource cost, and more significant values of $\beta$ do the opposite.

\section{Related Work}
\label{sec:relatedwork}

Configuration of machine learning inference systems has gained considerable attraction in recent years. Clipper~\cite{crankshaw2017clipper} is one of the early inference serving systems that introduced a general-purpose inference server with functionalities like caching, batching, and adaptive model selection.

MArk~\cite{zhang2019mark} employs request batching, predictive scaling, and serverless functions and proposes autoscaling policies that also take the hardware heterogeneity and service type diversity (FaaS, CaaS, IaaS) of inference serving data-centers into consideration.

INFaaS~\cite{romero2021infaas} provides an abstraction layer that decouples the model serving task from the used model for serving. Per each inference request, it searches through all the available sets of models for that specific inference task. Based on the request requirement, it finds the suitable model variant and dynamically offloads and unloads models as the user requirements change.

Model switching \cite{zhang2020model} is the first work that proposes switching between lightweight and heavier models as a workload adaptation mechanism. To be responsive to workload surges, it switches to a smaller but less accurate model. Unlike \namex, their model is not cost-aware and can only work under a fixed resource budget. 

Cocktail~\cite{gunasekaran2022cocktail} is the most similar work to \namex. It proposes an approach based on ensemble learning to reduce the cost while meeting the previous works' latency and accuracy efficiency. Cocktail uses ensembling as its accuracy maximization technique, which is costly as all the requests should be sent to all the ML models. Most of the time, many model sets should be used to get to the accuracy of the largest model. Cocktail uses transient virtual machines to improve cost efficiency. Nevertheless, using unstable transient instances can cause interruptions in the inference service. Deploying \namex on CaaS platforms like Google Autopilot does not suffer from similar problems. Due to fundamental structural differences and different problem formulations, we could not compare \namex with Cocktail.

\section{Future Works}
\label{sec:future}


\noindent\textbf{Hardware Heterogeneity.} While in this work, we focused on homogeneous CPU inferencing, the performance of \namex under general purposed (GPUs) and ASIC ML hardware can be evaluated. With packing requests into batches, GPUs can process higher workloads without a considerable increase in latency.

\noindent\textbf{Scalability with ML.} Our proposed solution to the ILP problem, works by brute-forcing through all possible configurations and picking the one that maximizes the objective function. Such an approach could suffer from scalability in case of growth in configuration space (more model variants e.g. scalability under one-for-all-networks \cite{cai2019once} with $10^{15}$ possible variants and bigger resource budgets in our case in a larger experimental setting). Utilizing ML-based solutions can decrease the amount of sampling in the search space, resulting in faster decision-making.

\noindent\textbf{Multi Model Serving.} In the case of using accelerators like GPUs, it is hard to share them among several containers, as there is no built-in mechanism for GPU sharing in container orchestration platforms like Kubernetes~\cite{gpu-sharing}. The multi-model deployment pattern, adapted in most production ML model servers~\cite{olston2017tensorflow, torchserve, tritonserver, seldon, kserve}, can mitigate the issues. Considering these emerging ML serving paradigms for improving adaptation mechanisms is a potential future work.


\vspace{-0.75em}
\section{Conclusion}
\label{sec:conclusion}
In this work, we presented \namex for ML inference services. It selects a set of ML model variants and their resource allocations to achieve a trade-off between accuracy and cost while preserving latency SLO guarantee. Experiments on real-world traces showed that \namex adapts better to dynamic workloads compared to the existing solutions by utilizing scaling and ML model variants selection.

\section*{Acknowledgements} This work has been supported in part by NSF (Awards 2233873, 2007202, and 2107463), German Research Foundation (DFG) within the Collaborative Research Center (CRC) 1053 MAKI, and Chameleon Cloud.

\bibliographystyle{ACM-Reference-Format}
\bibliography{ref}


\begin{thebibliography}{38}


\ifx \showCODEN    \undefined \def \showCODEN     #1{\unskip}     \fi
\ifx \showDOI      \undefined \def \showDOI       #1{#1}\fi
\ifx \showISBNx    \undefined \def \showISBNx     #1{\unskip}     \fi
\ifx \showISBNxiii \undefined \def \showISBNxiii  #1{\unskip}     \fi
\ifx \showISSN     \undefined \def \showISSN      #1{\unskip}     \fi
\ifx \showLCCN     \undefined \def \showLCCN      #1{\unskip}     \fi
\ifx \shownote     \undefined \def \shownote      #1{#1}          \fi
\ifx \showarticletitle \undefined \def \showarticletitle #1{#1}   \fi
\ifx \showURL      \undefined \def \showURL       {\relax}        \fi
\providecommand\bibfield[2]{#2}
\providecommand\bibinfo[2]{#2}
\providecommand\natexlab[1]{#1}
\providecommand\showeprint[2][]{arXiv:#2}

\bibitem[\protect\citeauthoryear{??}{gpu}{2022}]%
        {gpu-sharing}
 \bibinfo{year}{2022}\natexlab{}.
\newblock \bibinfo{title}{{GPU virtualization in K8S: challenges and state of
  the art}}.
\newblock
  \bibinfo{howpublished}{\url{https://www.arrikto.com/blog/gpu-virtualization-in-k8s-challenges-and-state-of-the-art/}}.
    (\bibinfo{date}{Nov} \bibinfo{year}{2022}).
\newblock


\bibitem[\protect\citeauthoryear{??}{hpa}{2022}]%
        {hpa}
 \bibinfo{year}{2022}\natexlab{}.
\newblock \bibinfo{title}{{Horizontal Pod autoscaling}}.
\newblock   (\bibinfo{date}{Jun} \bibinfo{year}{2022}).
\newblock
\showURL{%
\url{https://kubernetes.io/docs/tasks/run-application/horizontal-pod-autoscale/}}


\bibitem[\protect\citeauthoryear{??}{tf-}{2022a}]%
        {tf-inter}
 \bibinfo{year}{2022}\natexlab{a}.
\newblock \bibinfo{title}{{Inter-op parallelism threads}}.
\newblock   (\bibinfo{year}{2022}).
\newblock
\showURL{%
\url{https://www.tensorflow.org/api_docs/python/tf/config/threading/set_inter_op_parallelism_threads}}


\bibitem[\protect\citeauthoryear{??}{tf-}{2022b}]%
        {tf-intra}
 \bibinfo{year}{2022}\natexlab{b}.
\newblock \bibinfo{title}{{Intra-op parallelism threads}}.
\newblock   (\bibinfo{year}{2022}).
\newblock
\showURL{%
\url{https://www.tensorflow.org/api_docs/python/tf/config/threading/set_intra_op_parallelism_threads}}


\bibitem[\protect\citeauthoryear{??}{par}{2022}]%
        {parallelism_torch}
 \bibinfo{year}{2022}\natexlab{}.
\newblock \bibinfo{title}{{TorchServe parallelism threads}}.
\newblock   (\bibinfo{year}{2022}).
\newblock
\showURL{%
\url{https://pytorch.org/docs/stable/notes/cpu_threading_torchscript_inference.html}}


\bibitem[\protect\citeauthoryear{??}{kse}{2023}]%
        {kserve}
 \bibinfo{year}{2023}\natexlab{}.
\newblock \bibinfo{title}{{Kserve}}.
\newblock \bibinfo{howpublished}{\url{https://github.com/kserve/kserve}}.
  (\bibinfo{year}{2023}).
\newblock


\bibitem[\protect\citeauthoryear{??}{sel}{2023}]%
        {seldon}
 \bibinfo{year}{2023}\natexlab{}.
\newblock \bibinfo{title}{{Seldon}}.
\newblock
  \bibinfo{howpublished}{\url{https://github.com/SeldonIO/seldon-core}}.
  (\bibinfo{year}{2023}).
\newblock


\bibitem[\protect\citeauthoryear{??}{tri}{2023}]%
        {tritonserver}
 \bibinfo{year}{2023}\natexlab{}.
\newblock \bibinfo{title}{{Triton inference server}}.
\newblock
  \bibinfo{howpublished}{\url{https://github.com/triton-inference-server/server}}.
    (\bibinfo{year}{2023}).
\newblock


\bibitem[\protect\citeauthoryear{??}{vpa}{2023}]%
        {vpa}
 \bibinfo{year}{2023}\natexlab{}.
\newblock \bibinfo{title}{{Vertical Pod autoscaling}}.
\newblock
  \bibinfo{howpublished}{\url{https://github.com/kubernetes/autoscaler/tree/master/vertical-pod-autoscaler}}.
    (\bibinfo{year}{2023}).
\newblock


\bibitem[\protect\citeauthoryear{Akoush, Paleyes, Van~Looveren, and Cox}{Akoush
  et~al\mbox{.}}{2022}]%
        {akoush2022desiderata}
\bibfield{author}{\bibinfo{person}{Sherif Akoush}, \bibinfo{person}{Andrei
  Paleyes}, \bibinfo{person}{Arnaud Van~Looveren}, {and} \bibinfo{person}{Clive
  Cox}.} \bibinfo{year}{2022}\natexlab{}.
\newblock \showarticletitle{{Desiderata for next generation of ML model
  serving}}.
\newblock \bibinfo{journal}{{\em arXiv preprint arXiv:2210.14665\/}}
  (\bibinfo{year}{2022}).
\newblock


\bibitem[\protect\citeauthoryear{Amodei and Hernandez}{Amodei and
  Hernandez}{2019}]%
        {amodei-hernandez-2019}
\bibfield{author}{\bibinfo{person}{Dario Amodei} {and} \bibinfo{person}{Danny
  Hernandez}.} \bibinfo{year}{2019}\natexlab{}.
\newblock \bibinfo{title}{{AI and compute}}.
\newblock
  \bibinfo{howpublished}{\url{https://openai.com/blog/ai-and-compute/}}.
  (\bibinfo{date}{Nov} \bibinfo{year}{2019}).
\newblock


\bibitem[\protect\citeauthoryear{archiveteam}{archiveteam}{2021}]%
        {twitter-trace-2021-08}
\bibfield{author}{\bibinfo{person}{archiveteam}.}
  \bibinfo{year}{2021}\natexlab{}.
\newblock \bibinfo{title}{{Archiveteam-twitter-stream-2021-08}}.
\newblock
  \bibinfo{howpublished}{\url{https://archive.org/details/archiveteam-twitter-stream-2021-08}}.
    (\bibinfo{year}{2021}).
\newblock


\bibitem[\protect\citeauthoryear{Bar}{Bar}{2019}]%
        {bar2019}
\bibfield{author}{\bibinfo{person}{Jeff Bar}.} \bibinfo{year}{2019}\natexlab{}.
\newblock \bibinfo{title}{{Amazon EC2 ML inference}}.
\newblock \bibinfo{howpublished}{\url{https://tinyurl.com/5n8yb5ub}}.
  (\bibinfo{date}{Dec} \bibinfo{year}{2019}).
\newblock


\bibitem[\protect\citeauthoryear{Cai, Gan, Wang, Zhang, and Han}{Cai
  et~al\mbox{.}}{2019}]%
        {cai2019once}
\bibfield{author}{\bibinfo{person}{Han Cai}, \bibinfo{person}{Chuang Gan},
  \bibinfo{person}{Tianzhe Wang}, \bibinfo{person}{Zhekai Zhang}, {and}
  \bibinfo{person}{Song Han}.} \bibinfo{year}{2019}\natexlab{}.
\newblock \showarticletitle{{Once-for-all: train one network and specialize it
  for efficient deployment}}.
\newblock \bibinfo{journal}{{\em arXiv preprint arXiv:1908.09791\/}}
  (\bibinfo{year}{2019}).
\newblock


\bibitem[\protect\citeauthoryear{Contributors}{Contributors}{2020}]%
        {torchserve}
\bibfield{author}{\bibinfo{person}{PyTorch~Serve Contributors}.}
  \bibinfo{year}{2020}\natexlab{}.
\newblock \bibinfo{title}{{Torch serve}}.
\newblock \bibinfo{howpublished}{\url{https://pytorch.org/serve/}}.
  (\bibinfo{year}{2020}).
\newblock


\bibitem[\protect\citeauthoryear{Crankshaw, Wang, Zhou, Franklin, Gonzalez, and
  Stoica}{Crankshaw et~al\mbox{.}}{2017}]%
        {crankshaw2017clipper}
\bibfield{author}{\bibinfo{person}{Daniel Crankshaw}, \bibinfo{person}{Xin
  Wang}, \bibinfo{person}{Guilio Zhou}, \bibinfo{person}{Michael~J Franklin},
  \bibinfo{person}{Joseph~E Gonzalez}, {and} \bibinfo{person}{Ion Stoica}.}
  \bibinfo{year}{2017}\natexlab{}.
\newblock \showarticletitle{{Clipper: a low-latency online prediction serving
  system}}. In \bibinfo{booktitle}{{\em 14th $\{$USENIX$\}$ Symposium on
  Networked Systems Design and Implementation ($\{$NSDI$\}$ 17)}}.
  \bibinfo{pages}{613--627}.
\newblock


\bibitem[\protect\citeauthoryear{Gandhi, Harchol-Balter, Raghunathan, and
  Kozuch}{Gandhi et~al\mbox{.}}{2012}]%
        {gandhi2012autoscale}
\bibfield{author}{\bibinfo{person}{Anshul Gandhi}, \bibinfo{person}{Mor
  Harchol-Balter}, \bibinfo{person}{Ram Raghunathan}, {and}
  \bibinfo{person}{Michael~A Kozuch}.} \bibinfo{year}{2012}\natexlab{}.
\newblock \showarticletitle{{Autoscale: dynamic, robust capacity management for
  multi-tier data centers}}.
\newblock \bibinfo{journal}{{\em ACM Transactions on Computer Systems
  (TOCS)\/}} \bibinfo{volume}{30}, \bibinfo{number}{4} (\bibinfo{year}{2012}),
  \bibinfo{pages}{1--26}.
\newblock


\bibitem[\protect\citeauthoryear{Gujarati, Elnikety, He, McKinley, and
  Brandenburg}{Gujarati et~al\mbox{.}}{2017}]%
        {gujarati2017swayam}
\bibfield{author}{\bibinfo{person}{Arpan Gujarati}, \bibinfo{person}{Sameh
  Elnikety}, \bibinfo{person}{Yuxiong He}, \bibinfo{person}{Kathryn~S
  McKinley}, {and} \bibinfo{person}{Bj{\"{o}}rn~B Brandenburg}.}
  \bibinfo{year}{2017}\natexlab{}.
\newblock \showarticletitle{{Swayam: distributed autoscaling to meet SLAs of
  machine learning inference services with resource efficiency}}. In
  \bibinfo{booktitle}{{\em Proceedings of the 18th ACM/IFIP/USENIX Middleware
  Conference}}. \bibinfo{pages}{109--120}.
\newblock


\bibitem[\protect\citeauthoryear{Gujarati, Karimi, Alzayat, Hao, Kaufmann,
  Vigfusson, and Mace}{Gujarati et~al\mbox{.}}{2020}]%
        {gujarati2020serving}
\bibfield{author}{\bibinfo{person}{Arpan Gujarati}, \bibinfo{person}{Reza
  Karimi}, \bibinfo{person}{Safya Alzayat}, \bibinfo{person}{Wei Hao},
  \bibinfo{person}{Antoine Kaufmann}, \bibinfo{person}{Ymir Vigfusson}, {and}
  \bibinfo{person}{Jonathan Mace}.} \bibinfo{year}{2020}\natexlab{}.
\newblock \showarticletitle{{Serving DNNs like clockwork: performance
  predictability from the bottom up}}.
\newblock \bibinfo{journal}{{\em arXiv preprint arXiv:2006.02464\/}}
  (\bibinfo{year}{2020}).
\newblock


\bibitem[\protect\citeauthoryear{Gunasekaran, Mishra, Thinakaran, Sharma,
  Kandemir, and Das}{Gunasekaran et~al\mbox{.}}{2022}]%
        {gunasekaran2022cocktail}
\bibfield{author}{\bibinfo{person}{Jashwant~Raj Gunasekaran},
  \bibinfo{person}{Cyan~Subhra Mishra}, \bibinfo{person}{Prashanth Thinakaran},
  \bibinfo{person}{Bikash Sharma}, \bibinfo{person}{Mahmut~Taylan Kandemir},
  {and} \bibinfo{person}{Chita~R Das}.} \bibinfo{year}{2022}\natexlab{}.
\newblock \showarticletitle{Cocktail: a multidimensional optimization for model
  serving in cloud}. In \bibinfo{booktitle}{{\em USENIX NSDI}}.
  \bibinfo{pages}{1041--1057}.
\newblock


\bibitem[\protect\citeauthoryear{{Gurobi Optimization, LLC}}{{Gurobi
  Optimization, LLC}}{2023}]%
        {gurobi}
\bibfield{author}{\bibinfo{person}{{Gurobi Optimization, LLC}}.}
  \bibinfo{year}{2023}\natexlab{}.
\newblock \bibinfo{title}{{Gurobi optimizer reference manual}}.
\newblock   (\bibinfo{year}{2023}).
\newblock
\showURL{%
\url{https://www.gurobi.com}}


\bibitem[\protect\citeauthoryear{Hochreiter and Schmidhuber}{Hochreiter and
  Schmidhuber}{1997}]%
        {hochreiter1997long}
\bibfield{author}{\bibinfo{person}{Sepp Hochreiter} {and}
  \bibinfo{person}{J{\"u}rgen Schmidhuber}.} \bibinfo{year}{1997}\natexlab{}.
\newblock \showarticletitle{{Long short-term memory}}.
\newblock \bibinfo{journal}{{\em Neural computation\/}} \bibinfo{volume}{9},
  \bibinfo{number}{8} (\bibinfo{year}{1997}), \bibinfo{pages}{1735--1780}.
\newblock


\bibitem[\protect\citeauthoryear{Hu, Ghosh, and Govindan}{Hu
  et~al\mbox{.}}{2021}]%
        {hu2021scrooge}
\bibfield{author}{\bibinfo{person}{Yitao Hu}, \bibinfo{person}{Rajrup Ghosh},
  {and} \bibinfo{person}{Ramesh Govindan}.} \bibinfo{year}{2021}\natexlab{}.
\newblock \showarticletitle{{Scrooge: a cost-effective deep learning inference
  system}}. In \bibinfo{booktitle}{{\em Proceedings of the ACM Symposium on
  Cloud Computing}}. \bibinfo{pages}{624--638}.
\newblock


\bibitem[\protect\citeauthoryear{Keahey, Anderson, Zhen, Riteau, Ruth,
  Stanzione, Cevik, Colleran, Gunawi, Hammock, Mambretti, Barnes, Halbach,
  Rocha, and Stubbs}{Keahey et~al\mbox{.}}{2020}]%
        {keahey2020lessons}
\bibfield{author}{\bibinfo{person}{Kate Keahey}, \bibinfo{person}{Jason
  Anderson}, \bibinfo{person}{Zhuo Zhen}, \bibinfo{person}{Pierre Riteau},
  \bibinfo{person}{Paul Ruth}, \bibinfo{person}{Dan Stanzione},
  \bibinfo{person}{Mert Cevik}, \bibinfo{person}{Jacob Colleran},
  \bibinfo{person}{Haryadi~S. Gunawi}, \bibinfo{person}{Cody Hammock},
  \bibinfo{person}{Joe Mambretti}, \bibinfo{person}{Alexander Barnes},
  \bibinfo{person}{Fran\c{c}ois Halbach}, \bibinfo{person}{Alex Rocha}, {and}
  \bibinfo{person}{Joe Stubbs}.} \bibinfo{year}{2020}\natexlab{}.
\newblock \showarticletitle{{Lessons learned from the Chameleon testbed}}.
\newblock In \bibinfo{booktitle}{{\em Proceedings of the 2020 USENIX Annual
  Technical Conference (USENIX ATC '20)}}. \bibinfo{publisher}{USENIX
  Association}.
\newblock


\bibitem[\protect\citeauthoryear{Leopold}{Leopold}{2019}]%
        {leopold2019}
\bibfield{author}{\bibinfo{person}{George Leopold}.}
  \bibinfo{year}{2019}\natexlab{}.
\newblock \bibinfo{title}{{AWS to offer Nvidia’s T4 GPUs for AI
  inferencing}}.
\newblock
  \bibinfo{howpublished}{\url{https://www.hpcwire.com/2019/03/19/aws-upgrades-its-gpu-backed-ai-inference-platform/}}.
    (\bibinfo{date}{Mar} \bibinfo{year}{2019}).
\newblock


\bibitem[\protect\citeauthoryear{Nigade, Bauszat, Bal, and Wang}{Nigade
  et~al\mbox{.}}{2022}]%
        {nigade2022jellyfish}
\bibfield{author}{\bibinfo{person}{Vinod Nigade}, \bibinfo{person}{Pablo
  Bauszat}, \bibinfo{person}{Henri Bal}, {and} \bibinfo{person}{Lin Wang}.}
  \bibinfo{year}{2022}\natexlab{}.
\newblock \showarticletitle{{Jellyfish: timely inference serving for dynamic
  edge networks}}. In \bibinfo{booktitle}{{\em 2022 IEEE Real-Time Systems
  Symposium (RTSS)}}. IEEE, \bibinfo{pages}{277--290}.
\newblock


\bibitem[\protect\citeauthoryear{Olston, Fiedel, Gorovoy, Harmsen, Lao, Li,
  Rajashekhar, Ramesh, and Soyke}{Olston et~al\mbox{.}}{2017}]%
        {olston2017tensorflow}
\bibfield{author}{\bibinfo{person}{Christopher Olston}, \bibinfo{person}{Noah
  Fiedel}, \bibinfo{person}{Kiril Gorovoy}, \bibinfo{person}{Jeremiah Harmsen},
  \bibinfo{person}{Li Lao}, \bibinfo{person}{Fangwei Li}, \bibinfo{person}{Vinu
  Rajashekhar}, \bibinfo{person}{Sukriti Ramesh}, {and} \bibinfo{person}{Jordan
  Soyke}.} \bibinfo{year}{2017}\natexlab{}.
\newblock \showarticletitle{{Tensorflow-Serving: flexible, high-performance ML
  serving}}.
\newblock \bibinfo{journal}{{\em arXiv preprint arXiv:1712.06139\/}}
  (\bibinfo{year}{2017}).
\newblock


\bibitem[\protect\citeauthoryear{Park, Naumov, Basu, Deng, Kalaiah, Khudia,
  Law, Malani, Malevich, Nadathur, et~al\mbox{.}}{Park et~al\mbox{.}}{2018}]%
        {park2018deep}
\bibfield{author}{\bibinfo{person}{Jongsoo Park}, \bibinfo{person}{Maxim
  Naumov}, \bibinfo{person}{Protonu Basu}, \bibinfo{person}{Summer Deng},
  \bibinfo{person}{Aravind Kalaiah}, \bibinfo{person}{Daya Khudia},
  \bibinfo{person}{James Law}, \bibinfo{person}{Parth Malani},
  \bibinfo{person}{Andrey Malevich}, \bibinfo{person}{Satish Nadathur},
  {et~al\mbox{.}}} \bibinfo{year}{2018}\natexlab{}.
\newblock \showarticletitle{{Deep learning inference in Facebook data centers:
  characterization, performance optimizations and hardware implications}}.
\newblock \bibinfo{journal}{{\em arXiv preprint arXiv:1811.09886\/}}
  (\bibinfo{year}{2018}).
\newblock


\bibitem[\protect\citeauthoryear{Razavi, Luthra, Koldehofe, Mühlhäuser, and
  Wang}{Razavi et~al\mbox{.}}{2022}]%
        {razavi2022fa2}
\bibfield{author}{\bibinfo{person}{Kamran Razavi}, \bibinfo{person}{Manisha
  Luthra}, \bibinfo{person}{Boris Koldehofe}, \bibinfo{person}{Max
  Mühlhäuser}, {and} \bibinfo{person}{Lin Wang}.}
  \bibinfo{year}{2022}\natexlab{}.
\newblock \showarticletitle{{FA2: fast, accurate autoscaling for serving deep
  learning inference with SLA guarantees}}. In \bibinfo{booktitle}{{\em 2022
  IEEE 28th Real-Time and Embedded Technology and Applications Symposium
  (RTAS)}}. \bibinfo{pages}{146--159}.
\newblock
\showDOI{%
\url{https://doi.org/10.1109/RTAS54340.2022.00020}}


\bibitem[\protect\citeauthoryear{Romero, Li, Yadwadkar, and Kozyrakis}{Romero
  et~al\mbox{.}}{2021}]%
        {romero2021infaas}
\bibfield{author}{\bibinfo{person}{Francisco Romero}, \bibinfo{person}{Qian
  Li}, \bibinfo{person}{Neeraja~J Yadwadkar}, {and} \bibinfo{person}{Christos
  Kozyrakis}.} \bibinfo{year}{2021}\natexlab{}.
\newblock \showarticletitle{{$\{$INFaaS$\}$: automated model-less inference
  serving}}. In \bibinfo{booktitle}{{\em 2021 USENIX Annual Technical
  Conference (USENIX ATC 21)}}. \bibinfo{pages}{397--411}.
\newblock


\bibitem[\protect\citeauthoryear{Rzadca, Findeisen, Swiderski, Zych, Broniek,
  Kusmierek, Nowak, Strack, Witusowski, Hand, et~al\mbox{.}}{Rzadca
  et~al\mbox{.}}{2020}]%
        {rzadca2020autopilot}
\bibfield{author}{\bibinfo{person}{Krzysztof Rzadca}, \bibinfo{person}{Pawel
  Findeisen}, \bibinfo{person}{Jacek Swiderski}, \bibinfo{person}{Przemyslaw
  Zych}, \bibinfo{person}{Przemyslaw Broniek}, \bibinfo{person}{Jarek
  Kusmierek}, \bibinfo{person}{Pawel Nowak}, \bibinfo{person}{Beata Strack},
  \bibinfo{person}{Piotr Witusowski}, \bibinfo{person}{Steven Hand},
  {et~al\mbox{.}}} \bibinfo{year}{2020}\natexlab{}.
\newblock \showarticletitle{{Autopilot: workload autoscaling at Google}}. In
  \bibinfo{booktitle}{{\em Proceedings of the Fifteenth European Conference on
  Computer Systems}}. \bibinfo{pages}{1--16}.
\newblock


\bibitem[\protect\citeauthoryear{Sarwinda, Paradisa, Bustamam, and
  Anggia}{Sarwinda et~al\mbox{.}}{2021}]%
        {sarwinda2021deep}
\bibfield{author}{\bibinfo{person}{Devvi Sarwinda},
  \bibinfo{person}{Radifa~Hilya Paradisa}, \bibinfo{person}{Alhadi Bustamam},
  {and} \bibinfo{person}{Pinkie Anggia}.} \bibinfo{year}{2021}\natexlab{}.
\newblock \showarticletitle{{Deep learning in image classification using
  residual network (ResNet) variants for detection of colorectal cancer}}.
\newblock \bibinfo{journal}{{\em Procedia Computer Science\/}}
  \bibinfo{volume}{179} (\bibinfo{year}{2021}), \bibinfo{pages}{423--431}.
\newblock


\bibitem[\protect\citeauthoryear{Shen, Chen, Jin, Zhao, Kong, Philipose,
  Krishnamurthy, and Sundaram}{Shen et~al\mbox{.}}{2019}]%
        {shen2019nexus}
\bibfield{author}{\bibinfo{person}{Haichen Shen}, \bibinfo{person}{Lequn Chen},
  \bibinfo{person}{Yuchen Jin}, \bibinfo{person}{Liangyu Zhao},
  \bibinfo{person}{Bingyu Kong}, \bibinfo{person}{Matthai Philipose},
  \bibinfo{person}{Arvind Krishnamurthy}, {and} \bibinfo{person}{Ravi
  Sundaram}.} \bibinfo{year}{2019}\natexlab{}.
\newblock \showarticletitle{{Nexus: a GPU cluster engine for accelerating
  DNN-based video analysis}}. In \bibinfo{booktitle}{{\em Proceedings of the
  27th ACM Symposium on Operating Systems Principles}}.
  \bibinfo{pages}{322--337}.
\newblock


\bibitem[\protect\citeauthoryear{Velikovich, Williams, Scheiner, Aleksic,
  Moreno, and Riley}{Velikovich et~al\mbox{.}}{2018}]%
        {velikovich2018semantic}
\bibfield{author}{\bibinfo{person}{Leonid Velikovich}, \bibinfo{person}{Ian
  Williams}, \bibinfo{person}{Justin Scheiner}, \bibinfo{person}{Petar~S
  Aleksic}, \bibinfo{person}{Pedro~J Moreno}, {and} \bibinfo{person}{Michael
  Riley}.} \bibinfo{year}{2018}\natexlab{}.
\newblock \showarticletitle{{Semantic lattice processing in contextual
  automatic speech recognition for Google assistant}}. In
  \bibinfo{booktitle}{{\em Interspeech}}. \bibinfo{pages}{2222--2226}.
\newblock


\bibitem[\protect\citeauthoryear{Wang, Yang, Yu, Wang, Li, Sun, He, and
  Zhang}{Wang et~al\mbox{.}}{2021}]%
        {wang2021morphling}
\bibfield{author}{\bibinfo{person}{Luping Wang}, \bibinfo{person}{Lingyun
  Yang}, \bibinfo{person}{Yinghao Yu}, \bibinfo{person}{Wei Wang},
  \bibinfo{person}{Bo Li}, \bibinfo{person}{Xianchao Sun},
  \bibinfo{person}{Jian He}, {and} \bibinfo{person}{Liping Zhang}.}
  \bibinfo{year}{2021}\natexlab{}.
\newblock \showarticletitle{{Morphling: fast, near-optimal auto-configuration
  for cloud-native model serving}}. In \bibinfo{booktitle}{{\em Proceedings of
  the ACM Symposium on Cloud Computing}}. \bibinfo{pages}{639--653}.
\newblock


\bibitem[\protect\citeauthoryear{Zhang, Yu, Wang, and Yan}{Zhang
  et~al\mbox{.}}{2019}]%
        {zhang2019mark}
\bibfield{author}{\bibinfo{person}{Chengliang Zhang}, \bibinfo{person}{Minchen
  Yu}, \bibinfo{person}{Wei Wang}, {and} \bibinfo{person}{Feng Yan}.}
  \bibinfo{year}{2019}\natexlab{}.
\newblock \showarticletitle{{MArk: exploiting cloud services for
  cost-effective, SLO-aware machine learning inference serving}}. In
  \bibinfo{booktitle}{{\em 2019 $\{$USENIX$\}$ Annual Technical Conference
  ($\{$USENIX$\}$$\{$ATC$\}$ 19)}}. \bibinfo{pages}{1049--1062}.
\newblock


\bibitem[\protect\citeauthoryear{Zhang, Ananthanarayanan, Bodik, Philipose,
  Bahl, and Freedman}{Zhang et~al\mbox{.}}{2017}]%
        {zhang2017live}
\bibfield{author}{\bibinfo{person}{Haoyu Zhang}, \bibinfo{person}{Ganesh
  Ananthanarayanan}, \bibinfo{person}{Peter Bodik}, \bibinfo{person}{Matthai
  Philipose}, \bibinfo{person}{Paramvir Bahl}, {and} \bibinfo{person}{Michael~J
  Freedman}.} \bibinfo{year}{2017}\natexlab{}.
\newblock \showarticletitle{{Live video analytics at scale with approximation
  and $\{$delay-tolerance$\}$}}. In \bibinfo{booktitle}{{\em 14th USENIX
  Symposium on Networked Systems Design and Implementation (NSDI 17)}}.
  \bibinfo{pages}{377--392}.
\newblock


\bibitem[\protect\citeauthoryear{Zhang, Elnikety, Zarar, Gupta, and Garg}{Zhang
  et~al\mbox{.}}{2020}]%
        {zhang2020model}
\bibfield{author}{\bibinfo{person}{Jeff Zhang}, \bibinfo{person}{Sameh
  Elnikety}, \bibinfo{person}{Shuayb Zarar}, \bibinfo{person}{Atul Gupta},
  {and} \bibinfo{person}{Siddharth Garg}.} \bibinfo{year}{2020}\natexlab{}.
\newblock \showarticletitle{{Model-switching: dealing with fluctuating
  workloads in machine-learning-as-a-service systems}}. In
  \bibinfo{booktitle}{{\em 12th $\{$USENIX$\}$ Workshop on Hot Topics in Cloud
  Computing (HotCloud 20)}}.
\newblock


\end{thebibliography}
\clearpage
\section{Appendix}
\label{sec:appendix}

\begin{figure}[h]
\includegraphics[scale=0.55]{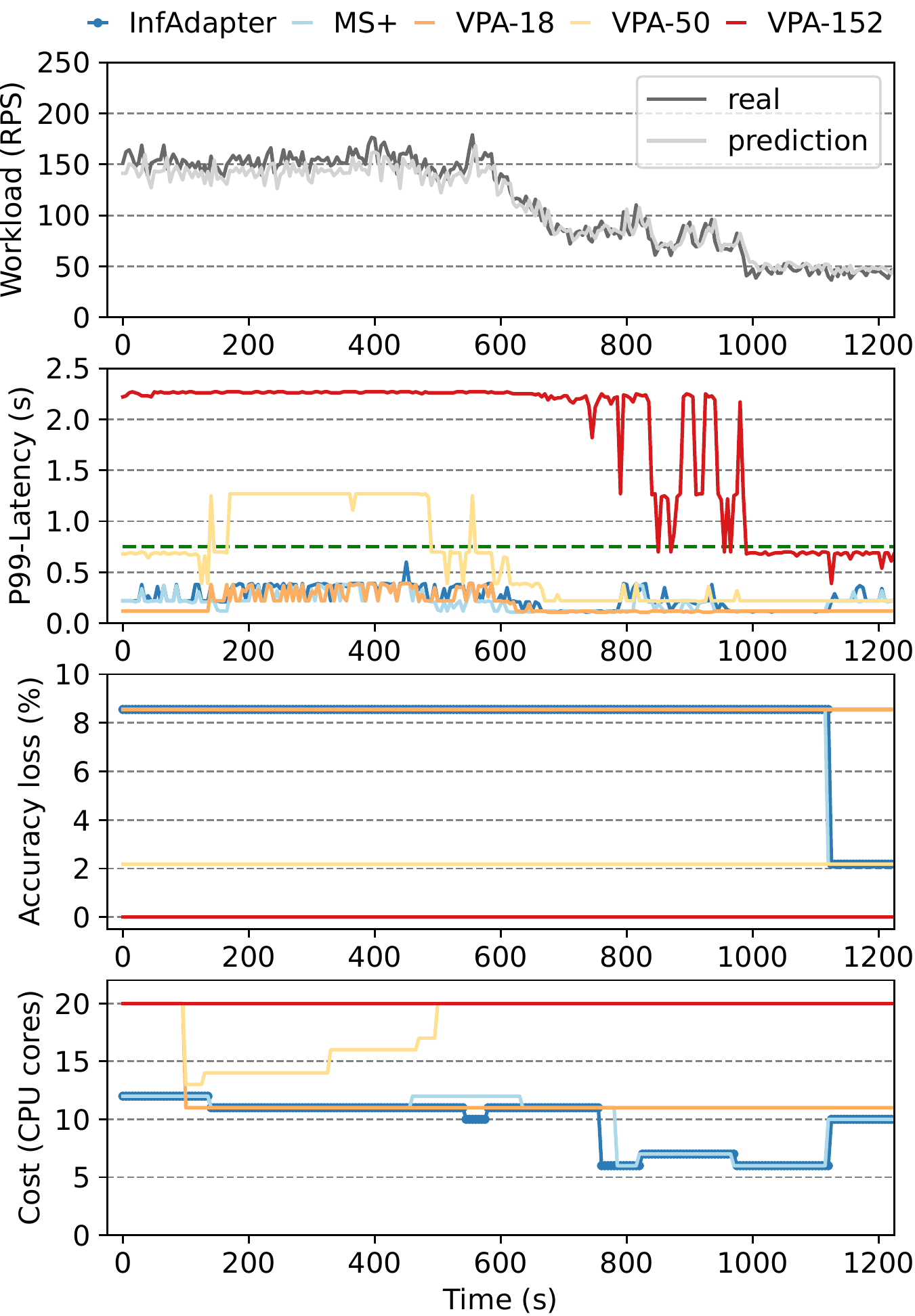}
\caption{Comparison of \namex with VPA used along with Resnet18, Resnet50 and Resnet152 on accuracy loss, cost and P99-latency, during the experiment with a non-bursty workload, $\beta$ = 0.2.}
\label{fig:comparison-0.2-nonbursty}
\end{figure}

\begin{figure}[h]
\includegraphics[scale=0.55]{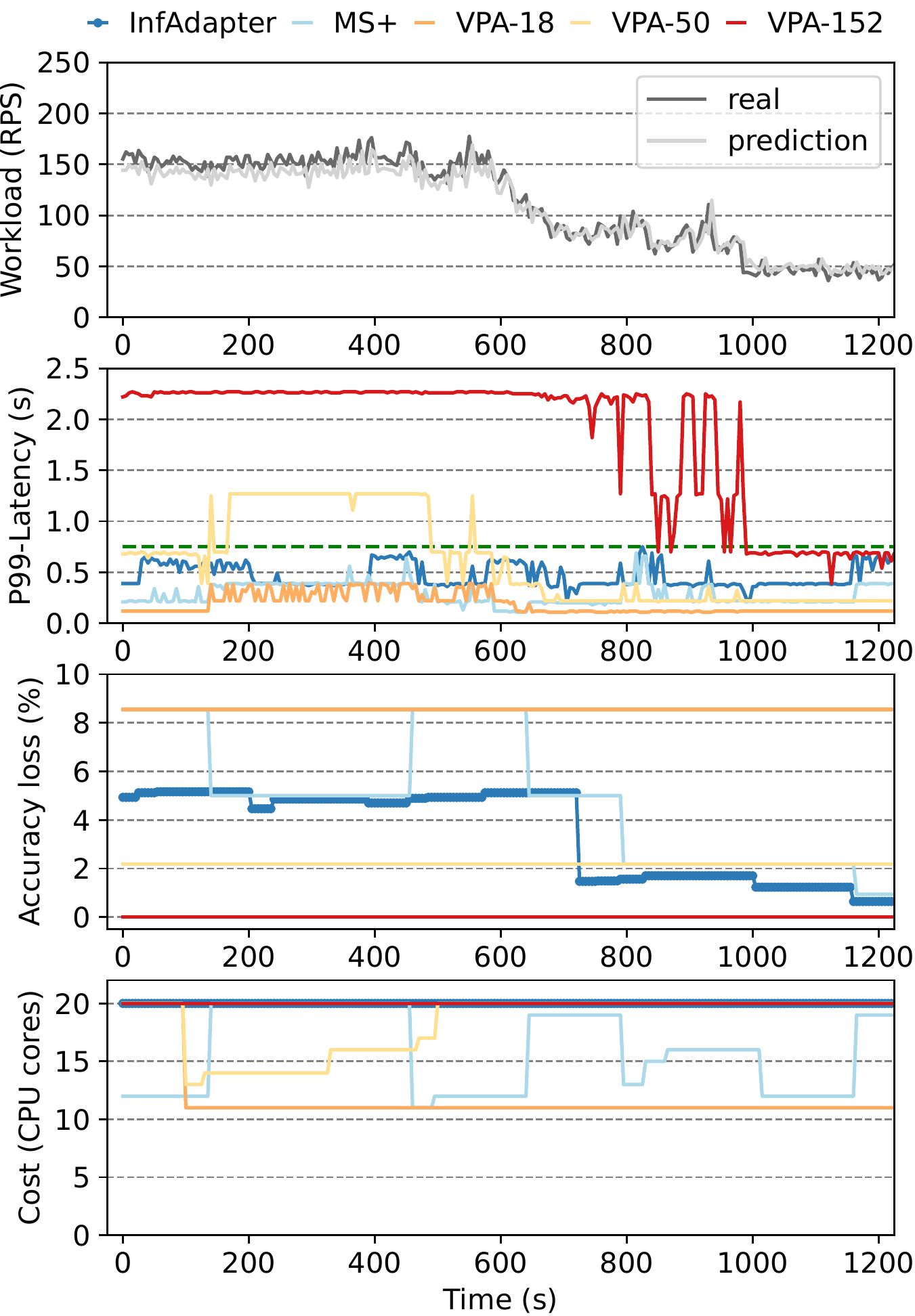}
\caption{Comparison of \namex with VPA used along with Resnet18, Resnet50 and Resnet152 on accuracy loss, cost and P99-latency, during the experiment with a non-bursty workload, $\beta$ = 0.0125.}
\label{fig:comparison-0.0125-nonbursty}
\end{figure}

$\beta$ introduced in Equation \ref{eq:ip} is the parameter responsible for increasing the effect of cost optimization in the optimization formulation. One of the main goals of \namex is to provide a tunable framework for achieving a trade-off between accuracy and cost. The relation between the values of $\alpha$ and $\beta$ is central to the behavior of \namex. Larger $\beta / \alpha$ ratios prioritize cost efficiency over accuracy optimization, and smaller values will result in the opposite. Figure \ref{fig:comparison-0.2-nonbursty} shows the case that \namex prioritize cost over cost optimization with the value $\beta = 0.2$ while in Figure \ref{fig:comparison-0.0125-nonbursty} a smaller value of $\beta = 0.0125$ results in achieving higher accuracy at the expense of sacrificing more cost.

\end{document}